# Connectionist Theory Refinement:
# Genetically Searching the Space of Network Topologies

**David W. Opitz**                                              OPITZ@CS.UMT.EDU
*Department of Computer Science*
*University of Montana*
*Missoula, MT 59812 USA*

**Jude W. Shavlik**                                            SHAVLIK@CS.WISC.EDU
*Computer Sciences Department*
*University of Wisconsin*
*1210 W. Dayton St.*
*Madison, WI 53706 USA*

## Abstract

An algorithm that learns from a set of examples should ideally be able to exploit the available resources of (a) abundant computing power and (b) domain-specific knowledge to improve its ability to generalize. Connectionist theory-refinement systems, which use background knowledge to select a neural network's topology and initial weights, have proven to be effective at exploiting domain-specific knowledge; however, most do not exploit available computing power. This weakness occurs because they lack the ability to refine the topology of the neural networks they produce, thereby limiting generalization, especially when given impoverished domain theories. We present the REGENT algorithm which uses (a) domain-specific knowledge to help create an initial *population* of knowledge-based neural networks and (b) genetic operators of crossover and mutation (specifically designed for knowledge-based networks) to continually search for better network topologies. Experiments on three real-world domains indicate that our new algorithm is able to significantly increase generalization compared to a standard connectionist theory-refinement system, as well as our previous algorithm for growing knowledge-based networks.

## 1. Introduction

Many scientific and industrial problems can be better understood by learning from samples of the task. For this reason, the machine learning and statistics communities devote considerable research effort to *inductive-learning* algorithms. Often, however, these learning algorithms fail to capitalize on a number of potentially available resources, such as domain-specific knowledge or computing power, that can improve their ability to generalize. Using domain-specific knowledge is desirable because inductive learners that start with an approximately correct theory can achieve improved "generalization" (accuracy on examples not seen during training) with significantly fewer training examples (Ginsberg, 1990; Ourston & Mooney, 1994; Pazzani & Kibler, 1992; Towell & Shavlik, 1994). Making effective use of available computing power is desirable because, for many applications, it is more important to obtain concepts that generalize well than it is to induce concepts quickly. In this article, we present an algorithm, called REGENT (REfining, with Genetic Evolution, Network Topologies), that utilizes available computer time to extensively search for a neural-network topology that best explains the training data while minimizing changes to a domain-specific theory.

Inductive learning systems that utilize a set of approximately correct, domain-specific inference rules (called a *domain theory*) which describe what is currently known about the domain, are called *theory-refinement* systems. Making use of this knowledge has been shown to be important since these rules may contain insight not easily obtainable from the current set of training examples (Ourston & Mooney, 1994; Pazzani & Kibler, 1992; Towell & Shavlik, 1994). For most domains, an expert who created the theory is willing to wait for weeks, or even months, if a learning system can produce an improved theory. Thus, given the rapid growth in computing power, we believe it is important for learning techniques to be able to trade off the expense of large numbers of computing cycles for gains in predictive accuracy. Analogous to *anytime planning* techniques (Dean & Boddy, 1988), we believe machine learning researchers should create better *anytime learning* algorithms. Such learning algorithms should produce a good concept quickly, then continue to search concept space, reporting the new "best" concept whenever one is found.

We concentrate on connectionist theory-refinement systems, since they have been shown to frequently generalize better than many other inductive-learning and theory-refinement systems (Fu, 1989; Lacher, Hruska, & Kuncicky, 1992; Towell, 1991). KBANN (Towell & Shavlik, 1994) is an example of such a connectionist system; it translates the provided domain theory into a neural network, thereby determining the network's topology, and then refines the reformulated rules using backpropagation (Rumelhart, Hinton, & Williams, 1986). However, KBANN, and other connectionist theory-refinement systems that do not alter their network topologies, suffer when given *impoverished* domain theories – ones that are missing rules needed to adequately learn the true concept (Opitz & Shavlik, 1995; Towell & Shavlik, 1994). TopGen (Opitz & Shavlik, 1995) is an improvement over these systems; it heuristically searches through the space of possible network topologies by adding hidden nodes to the neural representation of the domain theory. TopGen showed statistically significant improvements over KBANN in several real-world domains (Opitz & Shavlik, 1995); however, in this paper we empirically show that TopGen nevertheless suffers because it only considers simple expansions of the KBANN network.

To address this limitation, we broaden the types of topologies that TopGen considers by using genetic algorithms (GAs). We choose GAs for two reasons. First, GAs have been shown to be effective optimization techniques because of their efficient use of global information (Goldberg, 1989; Holland, 1975; Mitchell, 1996). Second, GAs have an inherent quality which makes them suitable for anytime learning. In "off-line" application mode (DeJong, 1975), GAs simulate many alternatives and output the best alternative seen so far.

Our new algorithm, REGENT, proceeds by first trying to generate, from the domain theory, a diversified initial population. It then produces new candidate networks via the genetic operators of *crossover* and *mutation*, after which these networks are trained using backpropagation. REGENT's crossover operator tries to maintain the rule structure of the network, while its mutation operator adds nodes to a network by using the TopGen algorithm. Hence, our genetic operators are specialized for connectionist theory refinement. Experiments reported herein show that REGENT is better able to search for network topologies than TopGen.

The rest of the paper is organized as follows. In the next section, we briefly argue for the importance of effectively exploiting data, theory, and available computer time in the learning process. We then review the KBANN and TopGen algorithms. We present the details of our REGENT algorithm in Section 4. This is followed by empirical results from three Human Genome Project domains. In Section 6, we discuss these results, as well as future work. We then review related work before concluding.

## 2. Using Data, Prior Knowledge, and Available CPU Cycles

A system that learns from a set of labeled examples is called an *inductive learner* (alternately, a *supervised*, *empirical*, or *similarity-based* learner). The output for each example is provided by a teacher, and the set of labeled examples given to a learner is called the *training* set. The task of inductive learning is to generate from the training set a concept description that correctly predicts the output of all future examples, not just those from the training set. Many inductive-learning algorithms have been previously studied (e.g., Michalski, 1983; Quinlan, 1986; Rumelhart et al., 1986). These algorithms differ both in their concept-representation language, and in their method (or *bias*) for constructing a concept within this language. These differences are important since they determine which concepts a classifier will induce.

An alternative to the inductive learning paradigm is to build a concept description not from a set of examples, but by querying experts in the field and directly assembling a set of rules that describe the concept (i.e., build an *expert system*; Waterman, 1986). A problem with building expert systems is that the theories derived from interviewing the experts tend to be only approximately correct. Thus, while the expert-provided domain theory is usually a good first approximation of the concept to be learned, inaccuracies are frequently exposed during empirical testing.

*Theory-refinement systems* (Ginsberg, 1990; Ourston & Mooney, 1994; Pazzani & Kibler, 1992; Towell & Shavlik, 1994) are systems that revise a theory on the basis of a collection of examples. These systems try to improve the theory by making minimal repairs to the theory to make it consistent with the training data. Changes to the initial domain theory should be kept to a minimum because the theory presumably contains useful information, even if it is not completely correct. These hybrid learning systems are designed to learn from both theory and data, and empirical tests have shown them to achieve high generalization with significantly fewer examples than purely inductive-learning techniques (Ourston & Mooney, 1994; Pazzani & Kibler, 1992; Towell & Shavlik, 1994). Thus, an ideal inductive-learning system should be able to incorporate any background knowledge that is available in the form of a domain theory to improve its ability to generalize.

As indicated earlier, available computer time is also an important resource since (a) computing power is rapidly increasing, and (b) for most problems an expert is willing to wait a lengthy period for an improved concept. For these reasons, one should develop "anytime" learning algorithms that can continually improve the quality of their answer over time. Dean and Boddy (1988) defined the criteria for an anytime algorithm to be: (a) the algorithm can be suspended and then resumed with minimal overhead, (b) the algorithm can be stopped at any time and return an answer, and (c) the algorithm must return answers that improve over time. While these criteria were created for planning and scheduling algorithms, they can apply to inductive learning algorithms as well.[1]

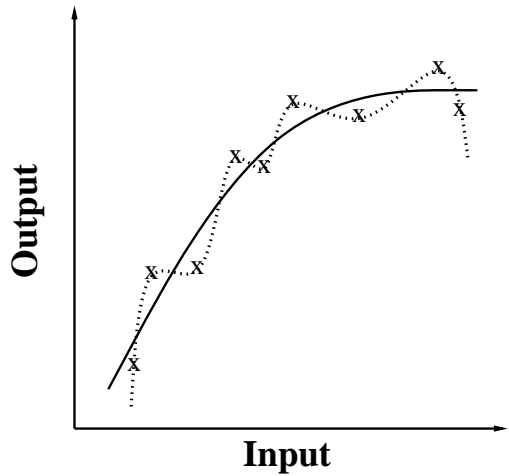

Figure 1: This is a classical regression example where a smooth function (the solid curve) that does not fit all of the noisy data points (the x's) is probably a better predictor than a high-degree polynomial (the dashed curve).

Most standard inductive learners, such as backpropagation (Rumelhart et al., 1986) and ID3 (Quinlan, 1986), are unable to continually improve their answers (at least until they receive additional training examples). In fact, if run too long, these algorithms tend to "overfit" the training set (Holder, 1991). Overfitting occurs when the learning algorithm produces a concept that captures too much information about the training examples, and not enough about the general characteristics of the domain as a whole. While these concepts do a great job of classifying the training instances, they do a poor job of generalizing to new examples – our ultimate measure of success. To help illustrate this point, consider the typical regression case shown in Figure 1. Here, fitting noisy data with a high-degree polynomial is likely to lead to poor generalization.

The general framework we use for encouraging our algorithm to improve its answer over time is quite simple. We spend our computer time considering many different possible concept descriptions, scoring each possibility, and always keeping the description that scores the best. Our framework is *anytime* with respect to the scoring function. The scoring function is only an *approximate* measure of generalization and is obviously still prone to the problems of overfitting; thus there is no guarantee that generalization will monotonically decrease over time. Nevertheless, assuming an *accurate* scoring function, then as long as we are considering a wide range of *good* possibilities, the quality of our best concept is likely to improve for a longer period of time.

---

1. Our use of the term *anytime learning* differs from that of Grefenstette and Ramsey (1992); they use it to mean continuous learning in a changing environment.

## 3. Review of KBANN and TopGen

The goal of this research is to exploit both prior knowledge and available computing cycles to search for the neural network that is most likely to generalize the best. We proceed by choosing, as an initial guess, the network defined by the KBANN algorithm. We then continually refine this topology to find the best network for our concept. Before presenting our new algorithm (REGENT), we give an overview of the KBANN algorithm as well as our initial approach of refining a KBANN-created network's topology (TopGen).

### 3.1 The KBANN Algorithm

KBANN (Towell & Shavlik, 1994) works by translating a domain theory consisting of a set of propositional rules directly into a neural network (see Figure 2). Figure 2a shows a Prolog-like rule set that defines membership in category $a$. Figure 2b represents the hierarchical structure of these rules, with solid lines representing necessary dependencies and dotted lines representing prohibitory dependencies. Figure 2c represents the network KBANN creates from this translation. It sets the biases so that nodes representing disjuncts have an output near 1 only when at least one of their high-weighted antecedents is satisfied, while nodes representing conjuncts must have all of their high-weighted antecedents satisfied (i.e., near 1 for positive links and near 0 for negative links). Otherwise activations are near 0. KBANN creates nodes $b1$ and $b2$ in Figure 2c to handle the two rules disjunctively defining $b$. The thin lines in Figure 2c represent low-weighted links that KBANN adds to allow these rules to add new antecedents during backpropagation training. Following network initialization, KBANN uses the available training instances to refine the network links. Refer to Towell (1991) or Towell and Shavlik (1994) for more details.

KBANN has been successfully applied to several real-world problems, such as the control of a chemical plant (Scott, Shavlik, & Ray, 1992), protein folding (Maclin & Shavlik, 1993), finding genes in a sequence of DNA (Opitz & Shavlik, 1995; Towell & Shavlik, 1994), and ECG patient monitoring (Watrous, Towell, & Glassman, 1993). In each case, KBANN was shown to produce improvements in generalization over standard neural networks for small numbers of training examples. In fact, Towell (1991) favorably compared KBANN with a wide variety of algorithms, including purely symbolic theory-refinement systems, on a version of the promoter and splice-junction tasks that we include as testbeds in Section 5.

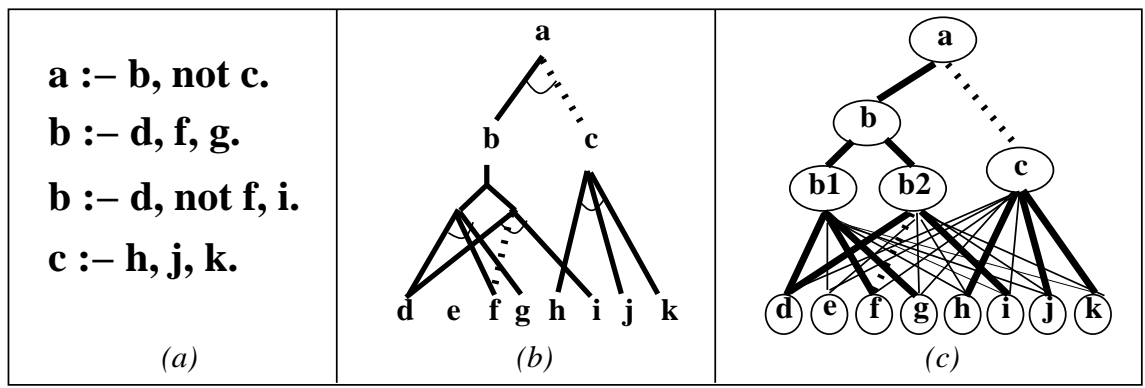

Figure 2: KBANN's translation of a knowledge base into a neural network. Panel (a) shows a sample propositional rule set in Prolog (Clocksin & Mellish, 1987) notation, panel (b) illustrates this rule set's corresponding AND/OR dependency tree, and panel (c) shows the resulting network created by KBANN's translation.

While training the KBANN-created network alters the antecedents of existing rules, it does not have the capability of inducing new rules because it does not add any additional hidden nodes during training. For instance, KBANN is unable to add a third rule for inferring $b$ in Figure 2's example. To help illustrate this point, consider the following example. Assume that Figure 2's target concept consists of Figure 2a's domain theory plus the rule:

$$b \text{ :- not } d, e, g.$$

Although we trained the KBANN network shown in Figure 2c with all possible examples of this target concept, it was unable to completely learn the conditions under which $a$ is true. The topology of the KBANN network must be modified in order to learn this new rule.

Studies show (Opitz & Shavlik, 1995; Towell, 1991) that while KBANN is effective at removing extraneous rules and antecedents in an expert-provided domain theory, its generalization ability suffers when given "impoverished" domain theories – theories that are missing rules or antecedents needed to adequately learn the true concept. *An ideal connectionist theory-refinement algorithm, therefore, should be able to dynamically expand the topology of its network during training.*

## 3.2 The TopGen Algorithm

TopGen (Opitz & Shavlik, 1995) addresses KBANN's limitation by heuristically searching through the space of possible expansions to a knowledge-based neural network – a network whose topology is determined by the direct mapping of the dependencies of a domain theory (e.g., a KBANN network). TopGen proceeds by first training the KBANN network, then placing it on a search queue. In each cycle, TopGen takes the best network from the search queue, estimates where errors occur in the network, adds new nodes in response to these estimates, trains these new networks, then places them back on the queue. TopGen judges where errors occur in a network by using training examples to increment two counters for each node, one for false negatives and one for false positives.

Figure 3 illustrates the possible ways TopGen can add nodes to one of its networks. In a symbolic rule base that uses negation-by-failure, one can decrease false negatives by either dropping antecedents from existing rules or adding new rules to the rule base. KBANN is effective at removing antecedents from existing rules, but is unable to add new rules; therefore, TopGen adds nodes, intended for decreasing false negatives, in a fashion that is analogous to adding a new rule to the rule base. If the existing node is an OR node, TopGen adds a new node as its child (see Figure 3a), and fully connects this new node to the input nodes. When the existing node is an AND node, TopGen creates a new OR node that is the parent of the original AND node and another new node that TopGen fully connects to the inputs (see Figure 3c); TopGen moves the outgoing links of the original node (A in Figure 3c) to become the outgoing links of the new OR node.

| Existing Node | Decrease False Negatives | Decrease False Positives |
|---|---|---|
| | | |

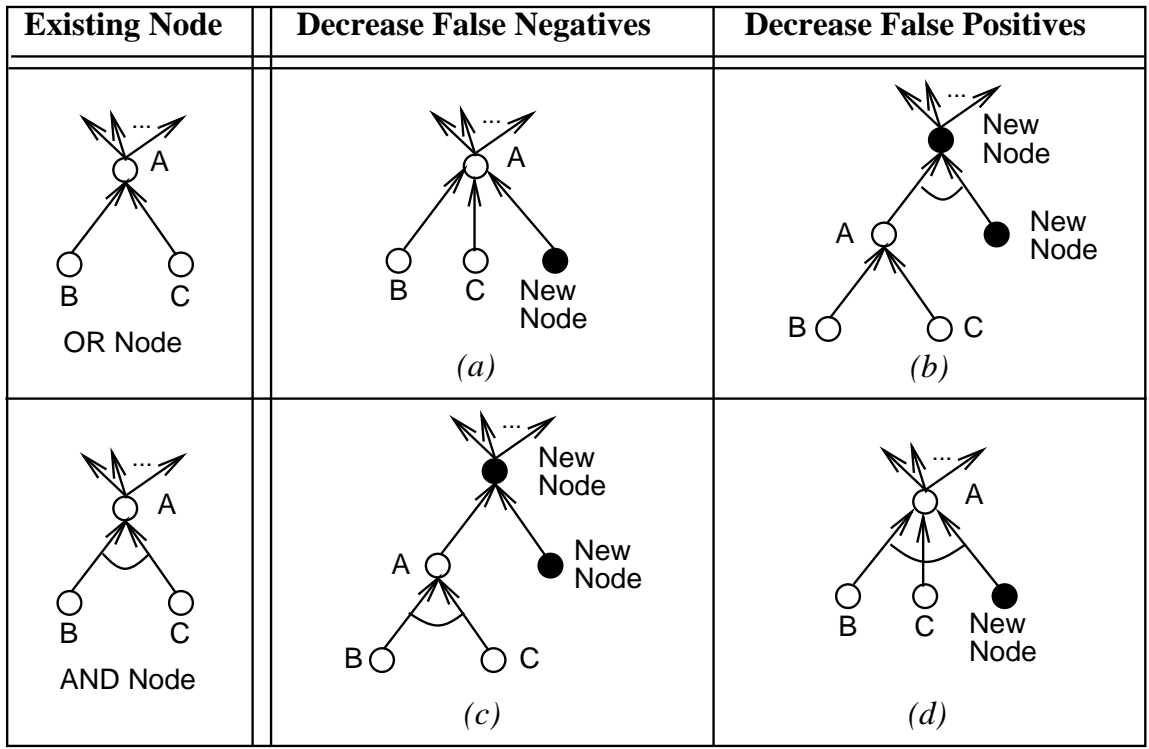

Figure 3: Possible ways to add new nodes to a knowledge-based neural network (arcs indicate AND nodes). To decrease false negatives, we wish to broaden the applicability of the node. Conversely, to decrease false positives, we wish to further constrain the node.

In a symbolic rule base, one can decrease false positives by either adding antecedents to existing rules or removing rules from the rule base. KBANN can effectively remove rules, but it is less effective at adding antecedents to rules and is unable to invent (i.e., constructively induce; Michalski, 1983) new terms as antecedents. Thus TopGen adds new nodes, intended to decrease false positives, in a fashion that is analogous to adding new constructively induced antecedents to the network. Figures 3b and 3d illustrates how this is done (analogous to Figures 3a and 3c explained above). Refer to Opitz and Shavlik (1993; 1995) for more details.

TopGen showed statistically significant improvements over KBANN in several real-world domains, and comparative experiments with a simpler approach to adding nodes verified that new nodes must be added in an intelligent manner (Opitz & Shavlik, 1995). In this article, however, we increase the number of networks TopGen considers during its search and show that the increase in generalization is primarily limited to the first few networks considered. Therefore, TopGen is not so much an "anytime" algorithm, but rather is a first step towards one. This is mostly due to the fact that TopGen only considers larger networks that contain the original KBANN network as subgraphs; however, as one increases the number of networks considered, one should also increase the variety of networks considered during the search. *Broadening the range of networks considered during the search through topology space is the major focus of this paper.*

## 4. The REGENT Algorithm

Our new algorithm, REGENT, tries to broaden the types of networks that TopGen considers with the use of GAs. We view REGENT as having two phases: (a) genetically searching through topology space, and (b) training each network using backpropagation's gradient descent method. REGENT uses the domain theory to aid in both phases. It uses the theory to help guide its search through topology space and to give a good starting point in weight space.

Table 1 summarizes the REGENT algorithm. REGENT first sets aside a *validation* set (from part of the *training* instances) for use in scoring the different networks. It then perturbs the KBANN-produced network to create an initial set of candidate networks. Next, REGENT trains these networks using backpropagation and places them into the population. In each cycle, REGENT creates new networks by crossing over and mutating networks from the current population that are randomly picked proportional to their fitness (i.e., validation-set correctness). It then trains these new networks and places them into the population. As it searches, REGENT keeps the network that has the lowest validation-set error as the best concept seen so far, breaking ties by choosing the smaller network in an application of Occam's Razor. A parallel version of REGENT trains many candidate networks at the same time using the Condor system (Litzkow, Livny, & Mutka, 1988), which runs jobs on idle workstations.

A diverse initial population will broaden the types of networks REGENT considers during its search; however, since the domain theory may provide useful information that may not be present in the training set, it is still desirable to use this theory when generating the initial population. REGENT creates diversity around the domain theory by randomly perturbing the KBANN network at various nodes. REGENT perturbs a node by either deleting it, or by adding new nodes to it in a manner analogous to one of TopGen's four methods for adding

---

**GOAL:** Search for the best network topology describing the domain theory and data.

1. Set aside a validation set from the training instances.

2. Perturb the KBANN-produced network in multiple ways to create initial networks, then train these networks using backpropagation and place them into the population.

3. Loop forever:

   (a) Create new networks using the crossover and mutation operators.

   (b) Train these networks with backpropagation, score with the validation set, and place into the population.

   (c) If a new network is the network with the lowest validation-set error seen so far (breaking ties by preferring the smallest network), report it as the current best concept.

Table 1: The REGENT algorithm.

**Crossover Two Networks:**

**GOAL:** Crossover two networks to generate two new network topologies.

1. Divide each network's hidden nodes into sets A and B using **DivideNodes**.

2. Set A forms one network, while set B forms another. Each new network is created as follows:

   (a) A network inherits weight $w_{ji}$ from its parent if nodes $i$ and $j$ either are also inherited or are input or output nodes.

   (b) Link unconnected nodes between levels with near-zero weights.

   (c) Adjust node biases to keep original AND or OR function of each node (see text for explanation).

**DivideNodes:**

**GOAL:** Divide the hidden nodes into sets A and B, while probabilistically maintaining each network's rule structure.

While some hidden node is not assigned to set A or B:

**(i)** Collect those unassigned hidden nodes whose output is linked only to either previously-assigned nodes or output nodes.

**(ii)** **If** set A or set B is empty:

    For each node collected in part (i), randomly assign it to set A or set B.

    **Else**

        Probabilistically add the nodes collected in part (i) to set A or set B. Equation 1 shows the probability of being assigned to set A. The probability of being assigned to set B is one minus this value.

Table 2: REGENT's method for crossing over networks.

---

nodes. (Should there happen to be multiple theories about a domain, all of them can be used to seed the population.)

## 4.1 REGENT's Crossover Operator

REGENT crosses over two networks by first dividing the nodes in each parent network into two sets, A and B, then combining the nodes in each set to form two new networks (i.e., the nodes in the two A sets form one network, while the nodes in the two B sets form another). Table 2 summarizes REGENT's method for crossover and Figure 4 illustrates it with an example. REGENT divides nodes, one level[2] at a time, starting at the level nearest the output nodes. When considering a level, if either set A or set B is empty, it cycles through each node in that level and randomly assigns it to either set. If neither set is empty, nodes are probabilistically placed into a set. The following equation calculates the probability of

---

2. Although one can define *level* several different ways, we define a node's *level* as the longest path from it to an output node.

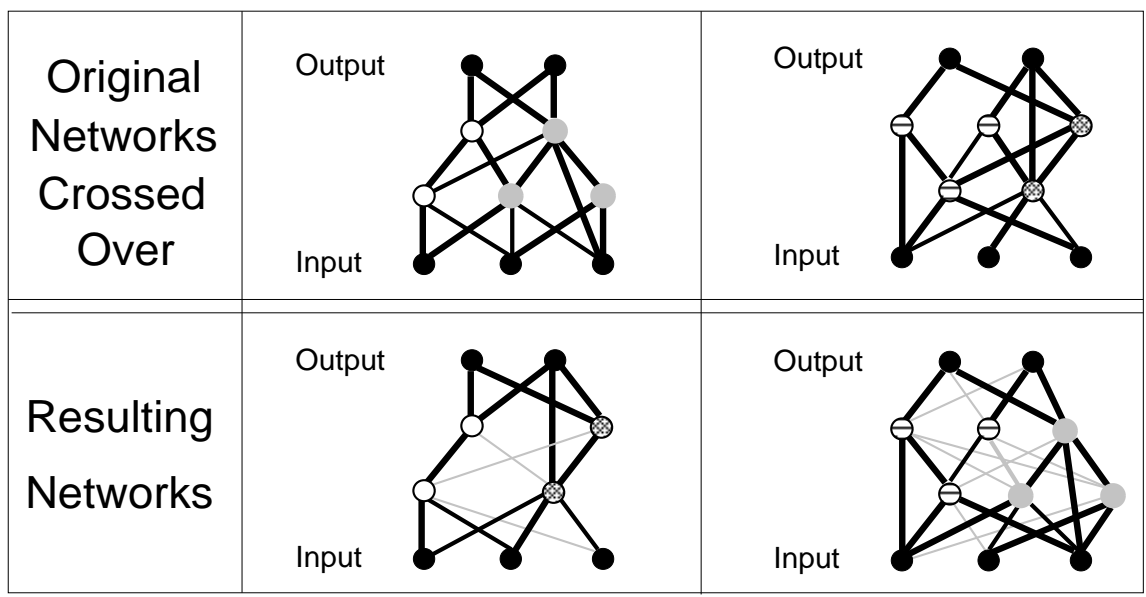

Figure 4: REGENT's method for crossing over two networks. The hidden nodes in each original network are divided into the sets A and B; the nodes in the two A sets form one new network, while the nodes in the two B sets form another. Grey lines represent low-weighted links that are added to fully connect neighboring levels.

a given node being assigned to set A:

$$Prob(node\ i \in set A) = \frac{\sum_{j \in A} |w_{ji}|}{\sum_{j \in A} |w_{ji}| + \sum_{j \in B} |w_{ji}|}, \tag{1}$$

where $j \in A$ means node $j$ is a member of set $A$ and $w_{ji}$ is the weight value from node $i$ to node $j$. The probability of belonging to set B is one minus this probability. With these probabilities, REGENT tends to assign to the same set those nodes that are heavily linked together. This helps to minimize the destruction of the rule structure of the crossed-over networks, since nodes belonging to the same syntactic rule are connected by heavily linked weights. Thus, REGENT's crossover operator produces new networks by crossing-over rules, rather than simply crossing-over nodes.

REGENT must next decide how to connect the nodes of the newly created networks. First, a new network inherits all weight values from its parents on links that (a) connect two nodes that are both inherited by the new network, (b) connect an inherited hidden node and an input or output node, or (c) directly connect an input node to an output node. It then adds randomly set, low-weighted links between unconnected nodes on consecutive levels.

Finally, it adjusts the bias of all AND or OR nodes to help maintain their *original* function. For instance, if REGENT removes a positively weighted incoming link for an AND node, it decrements the node's bias by subtracting the product of the link's magnitude and the average activation (over the set of training examples) entering that link. We do this because the bias for an AND node needs to be slightly less than the sum of the *positive* weights on the incoming links (see Towell and Shavlik, 1994 for more details). REGENT increments the bias for an OR node by an analogous amount when it removes negatively weighted incoming links (since the bias for an OR node should be slightly greater than the sum of the *negative* weights on the incoming links so that the node is inactive only when all incoming negatively weighted linked nodes are active and all positively weighted linked nodes are inactive).

## 4.2 REGENT's Mutation Operator

REGENT mutates networks by applying a variant of TopGen. REGENT uses TopGen's method for incrementing the false-negatives and false-positives counters for each node. REGENT then adds nodes, based on the values of these counters, the same way TopGen does. Since neural learning is effective at removing unwanted antecedents and rules from KNNs (see Section 3.1), REGENT only considers adding nodes, and not deleting them, during mutation. Thus, this mutation operator adds diversity to a population, while still maintaining a directed, heuristic-search technique for choosing where to add nodes; this directedness is necessary because we currently are unable to evaluate more than a few thousand possible networks per day.

## 4.3 Additional Details

REGENT adds newly trained networks to the population only if their validation-set correctness is better than or equal to an existing member of the population. When REGENT replaces a member, it replaces the member having the lowest correctness (ties are broken by choosing the oldest member). Other techniques (Goldberg, 1989), such as replacing the member nearest the new candidate network, can promote diverse populations; however, we do not want to promote diversity at the expense of decreased generalization. As a future research topic, we plan to investigate incorporating diversity-promoting techniques once we are able to consider tens of thousands of networks.

REGENT can be considered a Lamarckian[3], genetic-hillclimbing algorithm (Ackley, 1987), since it performs local optimizations on individuals, then passes the successful optimizations on to offspring. The ability of individuals to learn can smooth the fitness landscape and facilitate subsequent learning. Thus, Lamarckian learning can lead to a large increase in learning speed and solution quality (Ackley & Littman, 1994; Farmer & Belin, 1992).

## 5. Experimental Results

In this section, we test REGENT on three real-world Human Genome Project problems that aid in locating genes in DNA sequences (recognizing *promoters*, *splice-junctions*, and *ribosome-binding sites*). In these domains, the input is a short segment of DNA nucleotides (about 100 elements long) and the task is learn to predict if this DNA subsequence contains a biologically important site. Each domain is also accompanied by a domain theory generated by a DNA expert (M. Noordewier).

---

3. Lamarckian evolution is a theory based on the inheritance of characteristics acquired during a lifetime.

The promoter domain contains 234 positive examples, 702 negative examples, and 31 rules. The splice-junction domain contains 1,200 examples distributed equally among three classes, and 23 rules. Finally, the ribosome binding sites (RBS) domain, contains 366 positive examples, 1,098 negative examples, and 17 rules. (Note that the promoter data set and domain theory is a later version of the one that appears in Towell, 1994.) These domains are available at the University of Wisconsin Machine Learning (UW-ML) site via the World Wide Web (`ftp://ftp.cs.wisc.edu/machine-learning/shavlik-group/datasets/`) or anonymous ftp (`ftp.cs.wisc.edu`, then `machine-learning/shavlik-group/datasets`).

We first directly compare REGENT with TopGen and KBANN. We then perform a lesion study[4] on REGENT. In particular, we investigate the value of adding randomly created networks to REGENT's initial population and examine the utility of REGENT's genetic operators.

## 5.1 Experimental Methodology

All results in this article are from ten-fold cross validation runs. For each ten-fold cross validation the data set is first partitioned into ten equal-sized sets, then each set is in turn used as the test set while the classifier trains on the other nine sets. In each fold, REGENT is run with a population size of 20. Each network is trained using backpropagation. Parameter settings for the neural networks include a learning rate of 0.10, a momentum term of 0.9, and the number of training epochs of 20; the first two are standard settings and while 20 epochs may be fewer than typically found in the neural network literature, we set it at 20 to help avoid overfitting. We set aside a validation set consisting of 10% of the training examples for REGENT to use as its scoring function.

## 5.2 Generalization Ability of REGENT

This section's experiments compare the test-set accuracy (i.e., generalization) of REGENT with TopGen's. Figure 5 shows the test-set error of KBANN, TopGen, and REGENT as they search through the space of network topologies. The horizontal line in each graph results from the KBANN algorithm. We drew a horizontal line for the sake of visual comparison; recall that KBANN only considers a single network. The first point of each graph, after one network is considered, is nearly the same for all three algorithms, since they all start with the KBANN network; however, TopGen and REGENT differ slightly from KBANN since they must set aside part of the training set to score their candidate networks. Notice that TopGen stops improving after considering 10 to 30 networks and that the generalization ability of REGENT is better than TopGen's after this point. The reason for the occasional upward movements in Figure 5 is due to the fact that a validation set (or any scoring function) is an inexact estimate of the true generalization error (as are the results of the ten-fold cross validation).

Figure 6 presents the test-set error of TopGen and REGENT after they each consider 500 candidate topologies. The standard neural network results are from a fully connected, single-layer, feed-forward neural network; for each fold, we trained 20 networks containing up to 100 hidden nodes and used a validation set to choose the best network. Our results show KBANN generalizes much better than the best of these standard networks, thus further confirming KBANN's effectiveness in generating good network topologies. While TopGen is able to improve on the KBANN network, REGENT is able to significantly decrease the error rate over both KBANN and TopGen. (For benchmark purposes, REGENT has an error rate of 3.9% from a ten-fold cross validation on the full Splice Junction dataset of 3190 examples commonly used by machine learning researchers.)

---

4. A *lesion study* is one where components of an algorithm are individually disabled to ascertain their contribution to the full algorithm's performance (Kibler & Langley, 1988).

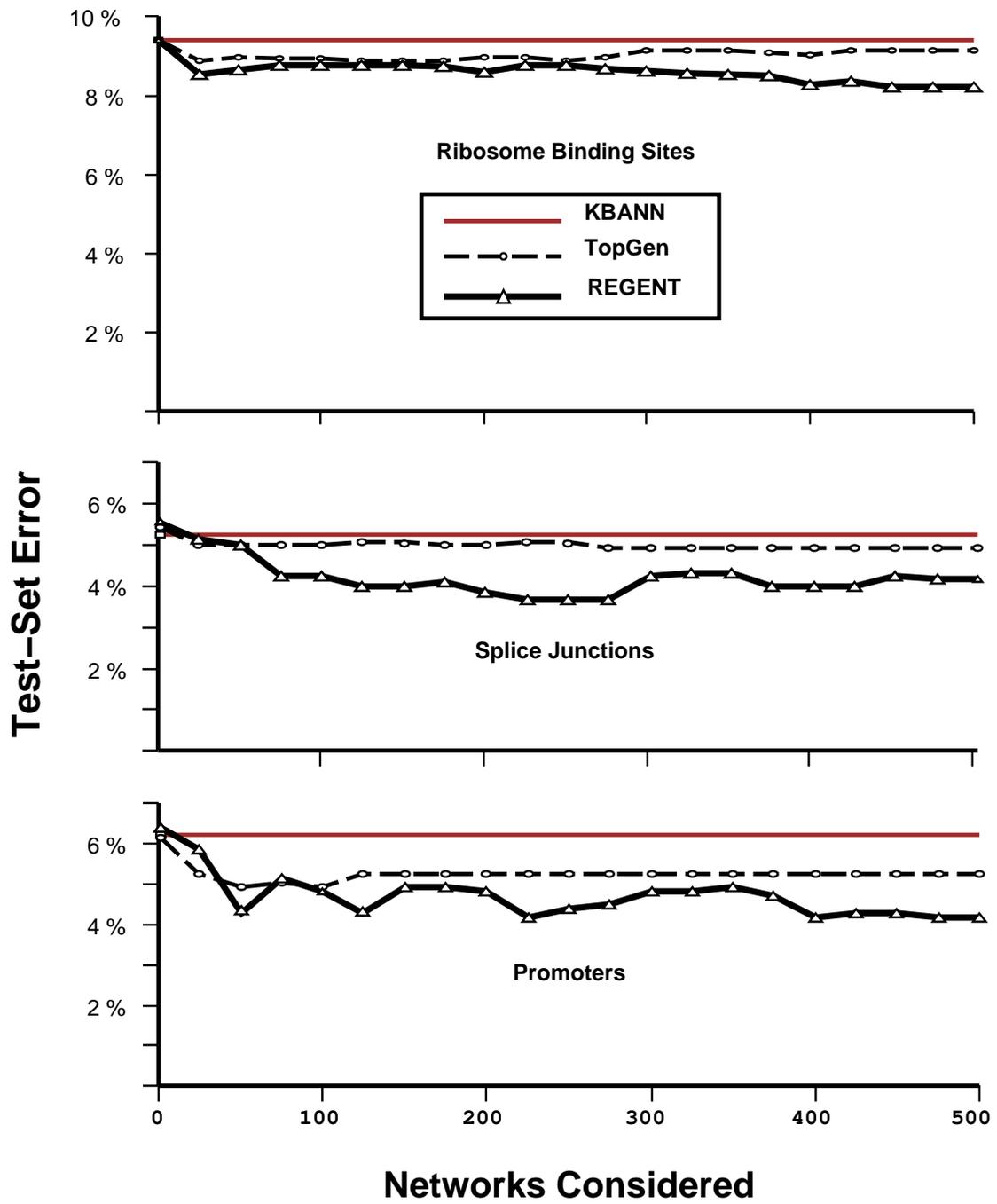

Figure 5: Error rates on the three Human Genome problems.

Table 3 contains the number of hidden nodes in the final networks produced by KBANN, TopGen, and REGENT. The results demonstrate that REGENT produces networks that are larger than both KBANN's and TopGen's networks (even though TopGen only *adds* nodes during its search). While REGENT's networks are larger, it does not necessarily mean that they are more "complex." We inspected sample networks and found that there are large portions of the network that are either not used (e.g., their weights are insignificantly small) or are functional duplications of other groups of hidden nodes.

One could prune weights and nodes during REGENT's search; however, such pruning can prematurely reduce the variety of structures available for recombination during crossover (Koza, 1992). Real-life organisms, for instance, have superfluous DNA that are believed to enhance the rate of evolution (Watson, Hopkins, Roberts, Argetsinger-Steitz, & Weiner, 1987). However, while pruning network size *during* genetic search may be unwise, one could prune REGENT's final network using, say, Hassibi and Stork's (1992) Optimal Brain Surgeon algorithm. This post-pruning process may increase the future classification speed of the network, as well as increase its comprehensibility and possibly its accuracy.

### 5.3 Lesion Study of REGENT

In this section, we describe a lesion study we performed on REGENT. Since a single run of REGENT takes about four CPU days to consider 500 networks, a single ten-fold cross

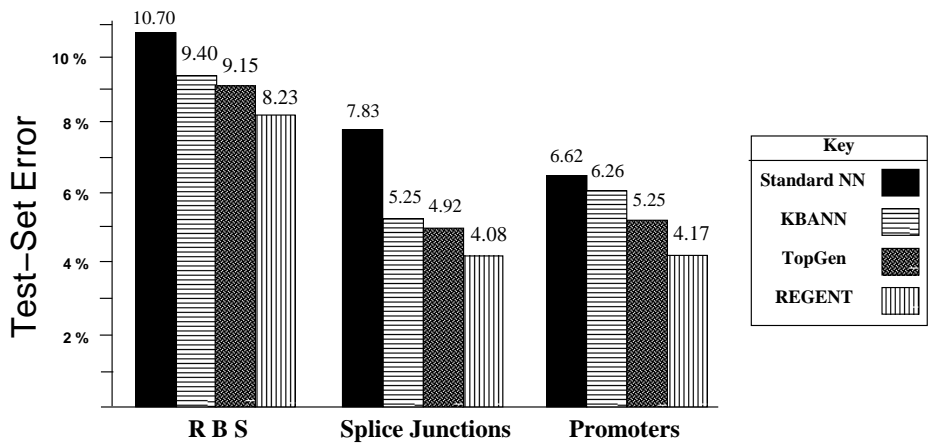

Figure 6: Test-set error rates after TopGen and REGENT each consider 500 networks. Pairwise, one-tailed *t*-tests indicate that REGENT differs from Standard NN, KBANN, and TopGen at the 95% confidence level on all three problems.

| Domain | KBANN | TopGen | REGENT |
|---|---|---|---|
| RBS | 18 | 42.1  (9.3) | 70.1  (25.1) |
| Splice Junction | 21 | 28.4  (4.1) | 32.4  (12.2) |
| Promoters | 31 | 40.2  (3.3) | 74.9  (38.9) |

Table 3: Number of hidden nodes in the networks produced by KBANN, TopGen, and REGENT. The columns show the mean number of hidden nodes found within these networks. Standard deviations are contained within parentheses; we do not report standard deviations for KBANN since it uses only one network.

---

validation takes (a minimum of) 40 CPU days. Therefore, given the inherent similarity of investigating various aspects of REGENT over multiple datasets, it is not feasible to run all experiments in this section until a 95% confidence level is reached in all cases (assuming that such a level actually exists). Nonetheless, these results convey important information about various components of REGENT, and, as shown in the previous section, the complete REGENT algorithm does generate statistically significant improvements over existing algorithms.

### 5.3.1 INCLUDING NON-KNNs IN REGENT'S POPULATION

The correct theory may be quite different from the initial domain theory. Thus, in this section we investigate whether one should include, in the initial population of networks, a variety of networks not obtained directly from the domain theory. Currently, REGENT creates its initial population by always perturbing the KBANN network. To include networks that are not obtained from the domain theory, we first randomly pick the number of hidden nodes to include in a network, then randomly create *all* of the hidden nodes in this network. We do this by adding new nodes to a randomly selected output or hidden node using one of TopGen's four methods for adding new nodes (refer to Figure 3). Adding nodes in this manner creates random networks whose node structure is analogous to dependencies found in symbolic rule bases, thus creating networks suitable for REGENT's crossover and mutation operators.

Table 4 shows the test-set error of REGENT with various percentages of knowledge-based neural networks (KNNs) present in the initial population. The first row contains the results of initializing REGENT with a purely random initial population (i.e., the population contains no KNNs). The second row lists the results when REGENT creates half its population with the domain theory, and the other half randomly. Finally, the last row contains the results of seeding the entire population with the domain theory.

These results suggest that including, in the initial population, networks that were not created from the domain theory increases REGENT's test-set error on all three domains. This occurs because the randomly generated networks are not as correct as the KNNs, and

|          | RBS  | Splice Junction | Promoters |
|----------|------|-----------------|-----------|
| 0% KNN   | 9.7% | 6.3%            | 5.1%      |
| 50% KNN  | 8.6% | 4.3%            | 4.6%      |
| 100% KNN | 8.2% | 4.1%            | 4.2%      |

Table 4: Test-set error after considering 500 networks. Each row gives the pergentage of KNNs present in the initial population. Pairwise, one-tailed $t$-tests indicate that initializing REGENT with 100% KNNs differs from 0% KNNs at the 95% confidence level on all three domains; however, the difference between the runs of 50% and 100% KNNs is not significant at this level.

---

thus offspring of the original KNN quickly replace the random networks. Hence, diversity in the population suffers compared to methods that start with a whole population of KNNs. Assuming the domain theory is not "malicious," it is therefore better to seed the entire population from the KBANN network. Should the domain theory indeed be malicious and contain information that promotes spurious correlations in the data, it would then be reasonable to randomly create the "whole" population. Running REGENT both with and without the domain theory allows one to investigate the utility of that theory.

These results are also interesting from a GA point of view. Forrest and Mitchell (1993) showed that GAs perform poorly on complex problems where the basic *building blocks* either (a) are non-trivial to find or (b) get split during crossover. Seeding the initial population with a domain theory (as REGENT does) can help define the basic building blocks for these problems.

### 5.3.2 Value of REGENT's Mutation

Typically with GAs, mutation is a secondary operation that is only sparingly used (Goldberg, 1989); however, REGENT's mutation is a directed approach that heuristically adds nodes to KNNs in a provenly effective manner (i.e., it uses TopGen). It is therefore reasonable to hypothesize that one should apply the mutation operator more frequently than traditionally done in GAs. The results in this section test this hypothesis.

Figure 7 presents the test-set error of REGENT with varying percentages of mutation (versus crossover) when creating new networks in step 3a of Table 1. Each graph plots four curves: (a) 0% mutation (i.e., REGENT only uses crossover), (b) 10% mutation, (c) 50% mutation, and (d) 100% mutation. Performing no mutations tests the value of solely using crossover, while 100% mutation tests the efficacy of the mutation operator by itself. Note that 100% mutation is just TopGen with a different search strategy; instead of keeping an OPEN list for heuristic search, a population of KNNs are generated and members of the population are improved proportional to their fitness. The other two curves (10% and 50% mutation) test the synergy between the two operators. Performing 10% mutation is closer to the traditional GA viewpoint that mutation is a secondary operation, while 50% mutation means that both operations are equally valuable. (Previous experiments in this section used 50% mutation and 50% crossover.)

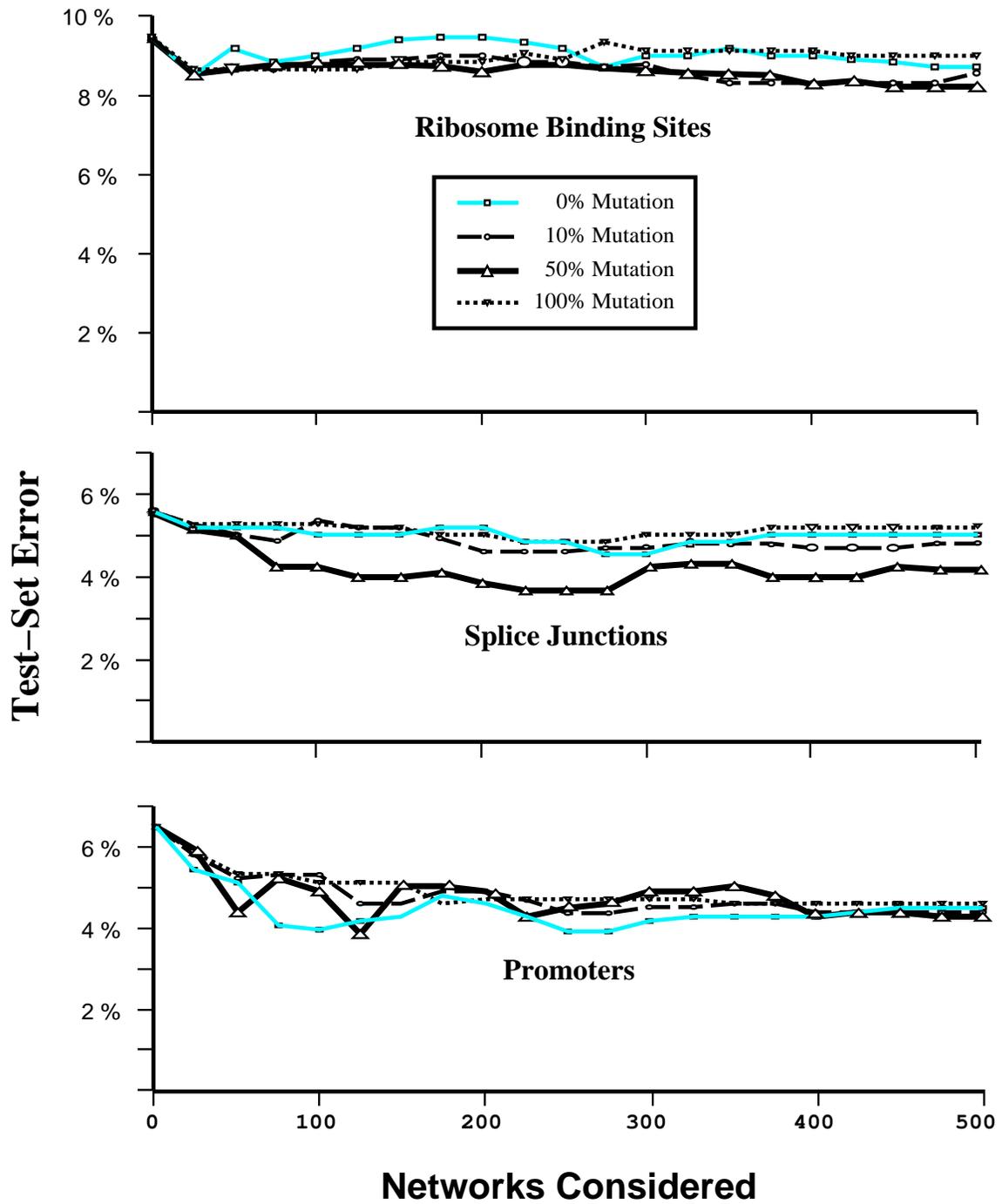

Figure 7: Error rates of REGENT with different fractions of mutation versus crossover after considering 500 networks. Arguably due to the inherent similarity of the algorithms, and the limited number of runs due to their computational complexity, the results are not significant at the 95% confidence level.

While the differences are not all statistically significant, the results nevertheless suggest that a synergy exists between the two operations. Except for the middle portion of the promoter domain, the results show that, qualitatively, using both operations at the same time is better than using either operation alone. In fact, equally mixing the mutation and crossover operator is better than the other three curves on all three domains once REGENT has considered 500 networks. This result is particularly pronounced on the splice-junction domain.

### 5.3.3 Value of REGENT's Crossover

REGENT tries to cross over the rules in the networks, rather than just blindly crossing over nodes. It does this by probabilistically dividing the nodes in the network into two sets where nodes belonging to the same rule tend to belong to the same set. In this section, we test the efficacy of REGENT's crossover by comparing it to a variant of itself where it *randomly* assigns nodes to two sets (rather than using **DivideNodes** in Table 2).

Table 5 contains the results of this test after 250 networks were considered. In the first row, REGENT-random-crossover, REGENT randomly breaks its hidden nodes into two sets, while in the second row, REGENT assigns nodes to two sets according to Table 2. In both cases, REGENT creates half its networks with its mutation operator, and the other half with crossover operator. Although the differences are not statistically significant, the results suggest that keeping the rule structure of the networks intact during crossover is important; otherwise, the basic building blocks of the networks (i.e., the rules) get split during crossover, and studies have shown the importance of keeping intact the basic building blocks during crossover (Forrest & Mitchell, 1993; Goldberg, 1989).

|  | Promoters | Splice Junction | RBS |
|---|---|---|---|
| REGENT-random-crossover | 4.6% | 4.7% | 9.1% |
| REGENT | 4.4% | 4.1% | 8.8% |

Table 5: Test-set error of two runs of REGENT: (a) randomly crossing over "nodes" in the networks, and (b) one with crossing over "rules" in the network (defined by Equation 1). Both runs considered 250 networks and used half crossover, half mutation. The results are not significant at the 95% confidence level; there is only a slight difference between the learning algorithms and the long run-times limited runs to a ten-fold cross validation.

## 6. Discussion and Future Work

Towell (1991) showed KBANN generalized better than many other machine learning algorithms on the promoter and splice-junction domains (the RBS dataset did not exist then).

Despite this success, REGENT is able to effectively use available computer cycles to significantly improve generalization over both KBANN and our previous improvement to KBANN, the TopGen algorithm. REGENT reduces KBANN's test-set error by 12% for the RBS domain, 22% for the splice-junction domain, and 33% for the promoter domain; it reduces TopGen's test-set error by 10% for the RBS domain, 17% for the splice-junction domain, and 21% for the promoter domain. Also, REGENT's ability to use available computing time is further aided by being inherently parallel, since we can train many networks simultaneously.

Further results show that REGENT's two genetic operators complement each other. The crossover operator considers a large variety of network topologies by probabilistically combining rules contained within two "successful" KNNs. Mutation, on the other hand, makes smaller, directed improvements to members of the population, while at the same time adding diversity to the population by adding new rules to the population. Equal use of both operators, therefore, allows a wide variety of topologies to be considered as well as allowing incremental improvements to members of the population.

Since REGENT searches through many candidate networks, it is important for it to be able to recognize the networks that are likely to generalize the best. With this in mind, our first planned extension of REGENT is to develop and test different network-evaluation functions. We currently use a validation set; however, validation sets have several drawbacks. First, keeping aside a validation set decreases the number of training instances available for each network. Second, the performance of a validation set can be a noisy approximator of the true error (MacKay, 1992; Weigend, Huberman, & Rumelhart, 1990). Finally, as we increase the number of networks searched, REGENT may start selecting networks that overfit the validation set. In fact, this explains the occasional upward trend in test-set error, from both TopGen and REGENT, in Figure 5.

To avoid the problem of overfitting the data, a common regression trick is to have a cost function that includes a "smoothness" term along with the error term. The best function, then, will be the smoothest function that also fits the data well. For neural networks, one can add to the estimated error a smoothness component that is a measure of the complexity of the network. The complexity of the network cannot simply be estimated by counting the number of possible parameters, since there tends to be significant duplication in the function of each weight in a network, especially early in the training process (Weigend, 1993). Two techniques that try to take into account the *effective* size of the network are Generalized Prediction Error (Moody, 1991) and Bayesian methods (MacKay, 1992).

Quinlan and Cameron-Jones (1995) propose adding an additional term to the accuracy and smoothness term that takes into account length of time spent searching. They coin the term "oversearching" to describe the phenomenon where more extensive searching causes lower predictive accuracy. Their claim is that oversearching is orthogonal to overfitting, and thus these complexity-based methods alone cannot prevent oversearching. As we increase the number of networks we consider during a search, we too may start oversearching, and thus plan to investigate adding an oversearching penalty term as well.

As indicated earlier, REGENT is Lamarckian in that it passes local optimizations of individuals (i.e., the trained weights of a network) to offspring. A viable alternative, called the Baldwin effect (Ackley & Littman, 1992; Baldwin, 1896; Belew & Mitchell, 1996; Hinton & Nowlan, 1987), is to have local search still change the fitness of an individual (backpropagation learning in this case), but then not pass these changes on to the offspring (this form of evolution is Darwinian in nature). Even though what is learned is not explicitly coded into the genetic material, individuals who are best able to learn will have the most offspring; thus learning still impacts evolution. In fact this form of evolution can sometimes outperform forms of Lamarckian evolution that employ the same local search strategy (Whitley, Gordon, & Mathias, 1994). Future work is to investigate the utility of the Baldwin effect on REGENT. In this case we would not cross over the trained networks, but instead cross over the initial weight settings *before* backpropagation learning took place.

Finally, often times there are multiple, even conflicting, theories about a domain. Future work, then, is to investigate ways of using all of these domain theories to seed the initial population. Although the results in Section 5.3.1 show that including randomly generated networks degrades generalization performance, seeding the population with multiple approximately correct theories should not degrade generalization, assuming the networks will have about the same initial correctness. Thus REGENT should be able to naturally combine good parts of multiple theories. Also, for a given domain theory, there are many different but equivalent ways to represent that theory using a set of propositional rules. Each representation leads to a different network topology, and even though each network starts with the same theory, some topologies may be more conducive to neural refinement.

## 7. Related Work

REGENT mainly differs from previous work in that it is an "anytime" theory-refinement system that continually searches, in a non-hillclimbing manner, for improvements to the domain theory. In summary, our work is unique in that it provides a connectionist approach that attempts to effectively utilize available background knowledge and available computer cycles to generate the best concept possible. We have broken the rest of this section into four parts: (a) connectionist theory-refinement algorithms, (b) purely symbolic theory-refinement algorithms, (c) algorithms that find an appropriate domain-specific neural-network topology, and (d) optimization algorithms wrapped around induction algorithms.

### 7.1 Connectionist Theory-Refinement Techniques

We begin our discussion with connectionist theory-refinement systems. These systems have been developed to refine many types of rule bases. For instance, a number of systems have been proposed for revising certainty-factor rule bases (Fu, 1989; Lacher et al., 1992; Mahoney & Mooney, 1993), finite-state automata (Maclin & Shavlik, 1993; Omlin & Giles, 1992), push-down automata (Das, Giles, & Sun, 1992), fuzzy-logic rules (Berenji, 1991; Masuoka, Watanabe, Kawamura, Owada, & Asakawa, 1990), and mathematical equations (Roscheisen, Hofmann, & Tresp, 1991; Scott et al., 1992). Most of these systems work like KBANN by first translating the domain knowledge into a neural network, then modifying the weights of this resulting network. Few attempts (which we describe next) have been made to dynamically adjust the resulting network's topology during training (as REGENT does).

Like both TopGen and REGENT, Fletcher and Obradovic (1993) present an approach that adds nodes to a KBANN network. Their system constructs a single layer of nodes, fully connected between the input and output nodes, "off to the side" of the KBANN network. They generate new hidden nodes using a variant of Baum and Lang's (1991) constructive algorithm. Baum and Lang's algorithm first divides the feature space with hyperplanes. They find each hyperplane by randomly selecting two points from different classes, then localizing a suitable split between these points. Baum and Lang repeat this process until they generate a fixed number of hyperplanes. Fletcher and Obradovic then map each of Baum and Lang's hyperplanes into one new hidden node, thus defining the weights between the input layer and that hidden node. Fletcher and Obradovic's algorithm does not change the weights of the Kbann portion of the network, so modifications to the initial rule base are solely left to the constructed hidden nodes. Thus, their system does not take advantage of Kbann's strength of removing unwanted antecedents and rules from the original rule base. In fact, TopGen compared favorably to a similar technique that also added nodes off to the side of Kbann (Opitz & Shavlik, 1993) and Regent outperformed TopGen in this article's experiments.

Rapture (Mahoney & Mooney, 1994) is designed for domain theories containing probabilistic rules. Like most connectionist theory-refinement systems, Rapture first translates the domain theory into a neural network, then refines the weights of the network with a modified backpropagation algorithm. Like Regent, Rapture is then able to dynamically refine the topology of its network. It does this by using the Upstart algorithm (Frean, 1990) to add new nodes to the network. Aside from being designed for probabilistic rules, Rapture differs from Regent in that it adds nodes with the intention of completely learning the training set, not generalizing well. Thus, while Rapture hillclimbs until the training set is learned, Regent continually searches topology space looking for a network that minimizes the scoring function's error. Also, Rapture initially only creates links that are specified in the domain theory, and only explicitly adds links through ID3's (Quinlan, 1986) information-gain metric. Regent, on the other hand, fully connect consecutive layers in their networks, allowing each rule the possibility of adding antecedents during training.

The Daid algorithm (Towell & Shavlik, 1992) is an extension to Kbann that uses the domain theory to help *train* the Kbann network. Since Kbann is more effective at dropping antecedents than adding them, Daid tries to find potentially useful inputs features not mentioned in the domain theory. It does this by backing-up errors to the lowest level of the domain theory, then computing correlations with the features. Daid then increases the weight of the links from the potentially useful input features based on these correlations. Daid mainly differs from Regent in that it does not refine the topology of the Kbann network. Thus, while Daid addresses Kbann's limitation of not effectively adding antecedents, it is still unable to introduce new rules or constructively induce new antecedents. Daid will therefore suffer with impoverished domain theories. Also notice that since Daid is an improvement for training KNNs, Regent can use Daid to train each network it considers during its search (however, we have not done so).

Opitz and Shavlik (1996) used a variant of Regent as their learning algorithm when generating a neural network "ensemble." A neural-network ensemble is a very successful technique where the outputs of a set of separately trained neural networks are combined to form one unified prediction (Drucker, Cortes, Jackel, LeCun, & Vapnik, 1994; Hansen & Salamon, 1990; Perrone, 1993). Since Regent considers many networks, it can select a subset of the final population of networks as an ensemble at minimal extra cost. Previous work, though, has shown that an ideal ensemble is one where the networks are both accurate and make their errors on different parts of the input space (Hansen & Salamon, 1990;

Krogh & Vedelsby, 1995). As a result, Opitz and Shavlik (1996) changed the scoring function of REGENT so that a "fit" network was now one that was both accurate and disagreed with the other members of the population as much as possible. In addition, their algorithm (ADDEMUP) actively tries to generate good candidates by emphasizing the current population's erroneous examples during backpropagation training. As a result of these alterations, ADDEMUP is able to create enough diversity among the population of networks to be able to effectively exploit the knowledge of the domain theory. Opitz and Shavlik (1996) show that ADDEMUP is able to generate a significantly better ensemble using the domain theory than either running ADDEMUP without the benefit of the theory or simply combining REGENT's final population of networks. Actively searching for a highly diverse population, however, does not aid in searching for the single best network. In fact, the single best network produced by ADDEMUP is significantly worse than REGENT's single best network on all three domains.

## 7.2 Purely Symbolic Theory-Refinement Techniques

Additional work related to REGENT includes purely symbolic theory-refinement systems that modify the domain theory directly in its initial form. Systems such as FOCL (Pazzani & Kibler, 1992) and FORTE (Richards & Mooney, 1995) are first-order, theory-refinement systems that revise predicate-logic theories. One drawback to these systems is that they currently do not generalize as well as connectionist approaches on many real-world problems, such as the DNA promoter task (Cohen, 1992).

There have been several genetic-based, first-order logic, multimodal concept learners (Greene & Smith, 1993; Janikow, 1993). Giordana and Saitta (1993) showed how to integrate one of these system, REGAL (Giordana, Saitta, & Zini, 1994; Neri & Saitta, 1996), with the deductive engine of ML-SMART (Bergadano, Giordana, & Ponsero, 1989) to help refine an incomplete or inconsistent domain theory. This version works by first using an automated theorem prover to recognize unresolved literals in a proof, then uses the GA-based REGAL to induce corrections to these literals. REGENT, on the other hand, use genetic algorithms (along with neural learning) to refine the *whole* domain theory at the same time.

DOGMA (Hekanaho, 1996) is a recently proposed GA-based learner that can use background knowledge to learn the same description language as REGAL. Current restrictions, however, force the representation language of the domain theory to be propositional rules. DOGMA converts a "flat" set of background rules (i.e., it does not handle intermediate conclusions) into individual bitstrings that are used as building blocks for a higher-level concept. DOGMA does not focus on theory refinement, rather it builds a completely new theory using substructures from the background knowledge. They term their approach as being more theory-suggested than theory-guided (Hekanaho, 1996).

Several systems, including ours, have been proposed for refining *propositional* rule bases. Early such approaches could only handle improvements to overly specific theories (Danyluk, 1989) or specializations to overly general theories (Flann & Dietterich, 1989). Later systems such as RTLS (Ginsberg, 1990), EITHER (Ourston & Mooney, 1994), PTR (Koppel, Feldman, & Segre, 1994), and TGCI (Donoho & Rendell, 1995) were later able to handle both types of refinements. We discuss the EITHER system as a representative of these propositional systems.

EITHER has four theory-revision operators: (a) removing antecedents from a rule, (b) adding antecedents to a rule, (c) removing rules from the rule base, and (d) inventing new rules. EITHER uses these operators to make revisions to the domain theory that correctly classify some of the previously misclassified training examples without undefining any of the correctly classified examples. EITHER uses inductive learning algorithms to invent new rules; it currently uses ID3 (Quinlan, 1986) as its induction component.

Even though REGENT's mutation operator add nodes in a manner analogous to how a symbolic system adds antecedents and rules, its underlying learning algorithm is "connectionist." Towell (1991) showed that KBANN outperformed EITHER on the promoter task, and REGENT outperformed KBANN in this article. KBANN's power on this domain is largely attributed to its ability to make "fine-grain" refinements to the domain theory (Towell, 1991). Because of EITHER's difficulty on this domain, Baffes and Mooney (1993) presented an extension to it called NEITHER-MOFN that is able to learn $M$-of-$N$ rules – rules that are true if $M$ of the $N$ antecedents are true. This improvement generated a concept that more closely matches KBANN's generalization performance.

While we want to minimize changes to a theory, we do not want to do it at the expense of accuracy; however, Donoho and Rendell (1995) demonstrate that most existing theory-refinement systems, such as EITHER, suffer in that they are only able to make small, local changes to the theory. Thus, when an accurate theory is significantly far in structure from the initial theory, these systems are forced to either become trapped in a local maximum similar to the initial theory, or are forced to drop entire rules and replace them with new rules that are inductively created purely from scratch. REGENT does not suffer from this in that it translates the theory into the less restricting representation of neural networks (Donoho & Rendell, 1995). Also, REGENT is able to further reconfigure the structure of the domain with genetic algorithms.

Many authors have reported results using varying subsets of the splice junction domain (e.g., Donoho and Rendell 1995; Mahoney 1996; Neri and Saitta 1996, and Towell and Shavlik 1994). While these authors used different training set sizes, it is nevertheless worthwhile to qualitatively discuss some of their conclusions here. Towell and Shavlik (1994) compared KBANN with numerous machine learning algorithms where each learning algorithm was given a training set of 1000 examples; KBANN's generalization ability compared favorably with these algorithms on the splice junction domain and REGENT, in turn, compared favorably with KBANN in this article. Donoho and Rendell (1995) showed their purely symbolic approach converged to the performance of KBANN at around 200 examples. Mahoney (1996) showed, using training set sizes of up to 400 examples, that his RAPTURE algorithm generalized better than KBANN on this domain; his results look similar to those of REGENT. Finally, Neri and Saitta (1996) showed that the generalization ability of the GA-based REGAL compares favorably to other purely symbolic, non-GA based techniques; while they used slightly different training set sizes than we did in this article, REGENT compares well to the results reported in their paper.

## 7.3 Finding Appropriate Network Topologies

Our third area of related work covers techniques that attempt to find a good domain-dependent topology by dynamically refining their network's topology during training. Many studies have shown that the generalization ability of a neural network depends on the topology of the network (Baum & Haussler, 1989; Tishby, Levin, & Solla, 1989). When trying to find an appropriate topology, one approach is to construct or modify a topology in an incremental fashion. *Network-shrinking* algorithms start with too many parameters, then remove nodes and weights during training (Hassibi & Stork, 1992; Le Cun, Denker, & Solla, 1989; Mozer & Smolensky, 1989). *Network-growing* algorithms, on the other hand, start with too few parameters, then add more nodes and weights during training (Blanziere & Katenkamp, 1996; Fahlman & Lebiere, 1989; Frean, 1990). The most obvious difference between REGENT and these algorithms is that REGENT uses domain knowledge and symbolic rule-refinement techniques to help determine the network's topology. Also, these other algorithms restructure their network based solely on training-set error, while REGENT minimized validation-set error.

Instead of incrementally finding an appropriate topology, one can mount a "richer" search than hillclimbing through the space of topologies. One common approach is to combine genetic algorithms and neural networks (as REGENT does). Genetic algorithms have been applied to neural networks in two different ways: (a) to optimize the connection weights in a fixed topology, and (b) to optimize the topology of the network. Techniques that solely use genetic algorithms to optimize weights (Montana & Davis, 1989; Whitley & Hanson, 1989) have performed competitively with gradient-based training algorithms; however, one problem with genetic algorithms is their inefficiency in fine-tuned local search, thus the scalability of these methods are in question (Yao, 1993). Kitano (1990b) presents a method that combines genetic algorithms with backpropagation. He does this by using the genetic algorithm to determine the starting weights for a network, which are then refined by backpropagation. REGENT differs from Kitano's method in that we use a domain theory to help determine each network's starting weights and genetically search, instead, for appropriate network topologies.

Most methods that use genetic algorithms to optimize a network topology are similar to REGENT in that they also use backpropagation to train each network's weights. Of these methods, many directly encode each link in the network (Miller, Todd, & Hegde, 1989; Oliker, Furst, & Maimon, 1992; Schiffmann, Joost, & Werner, 1992). These methods are relatively straightforward to implement, and are good at fine tuning small networks (Miller et al., 1989); however, they do not scale well since they require very large matrices to represent all the links in large networks (Yao, 1993). Other techniques (Dodd, 1990; Harp, Samad, & Guha, 1989; Kitano, 1990a) only encode the most important features of the network, such as the number of hidden layers, the number of hidden nodes at each layer, etc. These indirect encoding schemes can evolve different sets of parameters along with the network's topology and have been shown to have good scalability (Yao, 1993). Some techniques (Koza & Rice, 1991; Oliker et al., 1992) evolve both the architecture and connection weights at the same time; however, the combination of the two levels of evolution greatly increases the search space.

REGENT mainly differs from genetic-algorithm-based training methods in that it is designed for *knowledge-based* neural networks. Thus REGENT uses domain-specific knowledge and symbolic rule-refinement techniques to aid in determining the network's topology and initial weight setting. REGENT also differs in that it does not explicitly encode its networks; rather, in the spirit of Lamarkian evolution, it passes *trained* network weights to the off-spring. A final difference is that most of these other algorithms restructure their network based solely on training-set error, while REGENT minimizes validation-set error.

## 7.4 Wrapping Optimization Around Learning

We end our related work discussion with a brief overview of methods that combine global and local optimization strategies. Local search algorithms iteratively improve their estimate of the minimum by searching in only a local neighborhood of the current solution; local minima are not guaranteed to be global minima. (Many inductive learning methods are often equated with local optimization techniques; Rumelhart et al., 1986.) Global optimization methods (such as GAs), on the other hand, perform a more sophisticated search across multiple local minima and are good at finding regions of the search space where near-optimal solutions can be found; however, they are usually not as good at refining a solution (once it is close to a near-optimal solution) as local optimization strategies (Hart, 1994). Recent research has shown that it is desirable to emply both a global and local search strategy (Hart, 1994).

Hybrid GAs (such as REGENT) combine local search with a more traditional GA. While we focus on hybrid-GA algorithms in this section, this two-tiered search strategy has been employed by other researchers as well (Kohavi & John, 1997; Provost & Buchanan, 1995; Schaffer, 1993). GAs have been combined with many local search methods (Bala, Huang, Vafaie, DeJong, & Wechsler, 1995; Belew, 1990; Hinton & Nowlan, 1987; Turney, 1995). Neural networks are the most common choice for the local search strategy of hybrid GA systems and we discussed GA/neural-network hybrids in the Section 7.3. There are two common forms of hybrid GAs: Lamarckian-based evolution and Darwinian-based evolution (the Baldwin effect). Lamarckian evolution encodes its local improvements directly into its genetic material, while Darwinian evolution leaves the genetic material unchanged after learning. As discussed in Section 6, most authors use Lamarckian local search techniques and many have shown numerous cases where Lamarckian evolution outperforms non-Lamarckian local search (Belew, McInerney, & Schraudolph, 1992; Hart, 1994; Judson, Colvin, Meza, Huffa, & Gutierrez, 1992).

## 8. Conclusion

An ideal inductive-learning algorithm should be able to exploit the available resources of extensive computing power and domain-specific knowledge to improve its ability to generalize. KBANN (Towell & Shavlik, 1994) has been shown to be effective at translating a domain theory into a neural network; however, KBANN suffers in that it does not alter its topology. TopGen (Opitz & Shavlik, 1995) improved the KBANN algorithm by using available computer power to search for effective places to add nodes to the KBANN network; however, we show empirically that TopGen suffers from restricting its search to expansions of the KBANN network, and is unable to improve its performance after searching beyond a few topologies. Therefore TopGen is unable to exploit all available computing power to increase the correctness of an induced concept.

We present a new algorithm, REGENT, that uses a specialized genetic algorithm to broaden the types of topologies considered during TopGen's search. Experiments indicate that REGENT is able to significantly increase generalization over TopGen; hence, our new algorithm is successful in overcoming TopGen's limitation of only searching a small portion of the space of possible network topologies. In doing so, REGENT is able to generate a good solution quickly, by using KBANN, then is able to continually improve this solution as it searches concept space. Therefore, REGENT takes a step toward a true anytime theory refinement system that is able to make effective use of problem-specific knowledge and available computing cycles.

## Acknowledgements

This work was supported by Office of Naval Research grant N00014-93-1-0998 and National Science Foundation grant IRI 95-02990. Thanks to Richard Maclin, Richard Sutton, and three anonymous reviewers for their helpful comments. This is an extended version of a paper published in *Machine Learning: Proceedings of the Eleventh International Conference*, pp. 208-216, New Brunswick, NJ, Morgan Kaufmann. David Opitz completed a portion of this work while a graduate student at the University of Wisconsin and a professor at the University of Minnesota, Duluth.